\documentclass[runningheads,noend]{llncs}
\usepackage[T1]{fontenc}
\usepackage{graphicx}      
\usepackage{subcaption}    
\graphicspath{{figs/}}     
\usepackage[table]{xcolor}
\usepackage{makecell}

\definecolor{rowgray}{gray}{0.92}
\rowcolors{2}{rowgray}{white}

\makeatletter
\let\NAT@parse\undefined
\makeatother
\usepackage{hyperref}
\usepackage[utf8]{inputenc}
\usepackage{authblk}
\usepackage{bbm}
\usepackage{amsmath}
\usepackage[noend]{algpseudocode}  
\usepackage{marginnote}
\usepackage{amsfonts}
\usepackage{amsmath}
\usepackage{graphicx}
\usepackage[usenames,dvipsnames,table,xcdraw]{xcolor}
\usepackage{xspace}
\usepackage{caption}
\usepackage[table]{xcolor}
\usepackage{multirow}
\usepackage[usenames,dvipsnames,table,xcdraw]{xcolor}
\usepackage{colortbl}
\usepackage{multirow}
\usepackage{nicematrix}
\usepackage{makecell} 
\usepackage[ruled, linesnumbered, noend ]{algorithm2e}
\definecolor{lightblue}{rgb}{0.68, 0.85, 0.9}
\definecolor{lighterblue}{rgb}{0.80, 0.92, 0.95}
\definecolor{lightergray}{rgb}{0.90, 0.90, 0.90}
\definecolor{lighttan}{rgb}{0.82, 0.71, 0.55}
\definecolor{lightertan}{rgb}{0.94, 0.87, 0.80}
\definecolor{lightgreen}{rgb}{0.56, 0.93, 0.56}
\definecolor{lightergreen}{rgb}{0.74, 0.99, 0.79}
\definecolor{lightorange}{rgb}{1.00, 0.78, 0.49}
\definecolor{lighterorange}{rgb}{1.00, 0.88, 0.70}
\definecolor{lightred}{rgb}{1.0, 0.5, 0.5}
\definecolor{lighterred}{rgb}{1.0, 0.6, 0.6}
\usepackage[symbol]{footmisc}
\newcommand{\Term}[1]{\textsf{#1}}
\usepackage{subcaption}
\usepackage[nameinlink]{cleveref}
\definecolor{almostblack}{rgb}{0, 0, 0.3}
\newcommand{\FK}{\mathtt{FK}}
\hypersetup{%
      breaklinks,%
      colorlinks=true,%
      urlcolor=[rgb]{0.25,0.0,0.0},%
      linkcolor=[rgb]{0.5,0.0,0.0},%
      citecolor=[rgb]{0,0.2,0.445},%
      filecolor=[rgb]{0,0,0.4},
      anchorcolor=[rgb]={0.0,0.1,0.2}%
}

\usepackage{todonotes}
\newcommand{\R}{\mathbb{R}}

\newcommand{\1}{\mathbbm{1}}
\newcommand{\eps}{\varepsilon}

\newcommand{\DOF}{\Term{DOF}\xspace}

\newcommand{\cfg}{\mathtt{cfg}\xspace}

\newcommand{\etal}{\textit{et~al.}\xspace}

\newcommand{\HLink}[2]{\hyperref[#2]{#1~\ref*{#2}}}

\title{Serialized Red-Green-Gray: Quicker Heuristic Validation of Edges in Dynamic Roadmap Graphs}

\author{Yulie Arad\inst{1}\orcidID{0009-0005-8033-2631} \and
Stav Ashur\inst{1}\orcidID{0000-0003-0533-8978} \and
Marta Markowicz\inst{1}\orcidID{0009-0007-9486-4801} \and
James D. Motes\inst{1}\orcidID{0000-0002-9553-7331} \and
Marco Morales\inst{1,2}\orcidID{0000-0003-1824-2350} \and
Nancy M. Amato\inst{1}\orcidID{0000-0001-5817-5290}}

\authorrunning{Y. Arad et al.}
\institute{Siebel School of Computing and Data Science, 
University of Illinois, 201 N. Goodwin Avenue, Urbana, IL 61801, USA\\
\email{\{arad2,stava2,martasm2,jmotes2,namato\}@illinois.edu}
\and
Department of Computer Science, Instituto Tecnológico Autónomo de México (ITAM), Mexico City, 01080, México\
}

\begin{document}

\maketitle
\thispagestyle{empty}
\pagestyle{empty}

\begin{abstract}
Motion planning in dynamic environments, such as robotic warehouses, requires fast adaptation to frequent changes in obstacle poses. Traditional roadmap-based methods struggle in such settings, relying on inefficient reconstruction of a roadmap or expensive collision detection to update the existing roadmap. To address these challenges we introduce the \textbf{Red–Green–Gray (RGG)} framework, a method that builds on SPITE to quickly classify roadmap edges as invalid (red), valid (green), or uncertain (gray) using conservative geometric approximations. \textbf{Serial RGG} provides a high-performance variant leveraging batch serialization and vectorization to enable efficient GPU acceleration. Empirical results demonstrate that while RGG effectively reduces the number of unknown edges requiring full validation, SerRGG achieves a 2–9x speedup compared to the sequential implementation. This combination of geometric precision and computational speed makes SerRGG highly effective for time-critical robotic applications.

\end{abstract}

\section{Introduction}



Real-world robotic solutions must often account for changes in the environment such as  shifting warehouse configurations or unforeseen human movement which require the system to rapidly update its motion plan.
Motion planning roadmaps offer an efficient method for replanning by encoding feasible motions in a queryable graph for a robot within a known environment.
However, in dynamic environments, where obstacles may be moved, added, or removed, these precomputed roadmaps can quickly become outdated.
Even minor local changes can invalidate critical portions of the graph, making it unsafe for navigation.
A na\"{i}ve solution is to rebuild the roadmap from scratch after every change, but this is highly inefficient, since most edges typically do not change their validity status.
A more practical strategy is to update the roadmap incrementally, revalidating only those components whose status is uncertain.
To be effective in real-world applications, such revalidation must occur in real-time, making exhaustive collision checking impractical and motivating the need for methods that update roadmaps both selectively and efficiently.

Several approaches have been proposed to address the challenge of efficiently maintaining roadmaps in dynamic environments. The Dynamic Roadmap (DRM) method \cite{lh-frtppce-02} partitions the workspace into a grid and maps roadmap components (nodes and edges) to the workspace cells they intersect. When an obstacle moves, only the affected cells need to be checked, allowing for faster updates. However, DRM's effectiveness is highly sensitive to grid resolution: fine grids improve precision but require long preprocessing times and scale poorly in memory, while coarse grids lead to excessive conservatism and large numbers of collision checks for uncertain edges. More recently, SPITE \cite{almmmhpa-spitme-24} introduced the use of swept-volume approximations, over-approximating robot motions with geometric primitives known as cigars. Instead of relying on a fixed spatial discretization, SPITE organizes these approximations within a hierarchical axis-aligned bounding box (AABB) tree, enabling efficient broadphase filtering that scales to large, densely connected roadmaps. This approach can quickly validate many edges but, like DRM, it is conservative and often leaves a substantial number of edges unresolved. As the number of uncertain edges grows, the cost of revalidation can quickly dominate planning time, especially in large, densely connected roadmaps.

To address this we introduce the Red-Green-Gray (RGG) framework (preliminary version presented as a workshop paper \cite{arad2026heuristicedges}), an effective structure for maintaining and updating roadmaps in dynamic environments while still providing accurate geometric reasoning about edge validities. We augment the SPITE framework by using OBBs for more efficient over-approximations and introducing splines as under-approximations for the swept volumes. This dual-approximation approach enables edges to be classified in one of three categories: \textit{green} (guaranteed valid), \textit{red} (guaranteed invalid), and \textit{gray} (uncertain). By definitively resolving edges where obstacles either miss the over-approximation or intersect the under-approximation significantly, we significantly reduce redundant checks  across updates and focus computational resources on ambiguous gray edges.

However, while still a significant improvement over predecessors, on very large roadmaps with large numbers of revalidations, collision checking of the geometric primitives can still become slow. Thus we also present Serialized Red-Green-Gray (SerRGG), a serialized and vectorized realization of the RGG framework focused on improving computational speed rather than geometric accuracy.
Serialization enables the roadmap data to be stored and reloaded efficiently between planning iterations, while vectorization leverages parallel operations to accelerate edge validation and state updates.
SerRGG also introduces a GPU-friendly spatial subdivision strategy that aligns spatial filtering with batched, fixed capacity parallel computation, yielding a substantially faster implementation than all-pairs checks. Overall, SerRGG preserves the correctness and structure of the RGG framework while providing a faster implementation for real-time applications. 

In particular, we make the following contributions: 
\begin{enumerate}
    \item Red-Green-Gray Framework: Allows for quick detection of valid (green) and invalid (red) edges to minimize unknown validities (gray). Employs geometric primitives that are more precise representations than the original SPITE and also well suited for serialization
    \item Performance: SerRGG achieves substantial runtime improvements over SPITE implementations by using RGG framework and exploiting vectorized computation and optimized data handling.
    \item Spatially Filtered GPU Validation: SerRGG replaces the tree-based edge checking approach of sequential SPITE with a grid-based, spatially filtered validation strategy designed for GPU execution. By partitioning the workspace into fixed-capacity cells,
    SerRGG avoids global all-pairs enumeration while retaining high parallel throughput.
    \item Empirical Validation: We present a comprehensive experimental evaluation demonstrating RGG's and SerRGG's speedups and accuracy across a range of dynamic environments and benchmark scenarios. 


\end{enumerate}

\begin{figure}[t]
    \centering
    \subfloat[Fixed-base manipulator]{
        \includegraphics[width=0.45\linewidth]{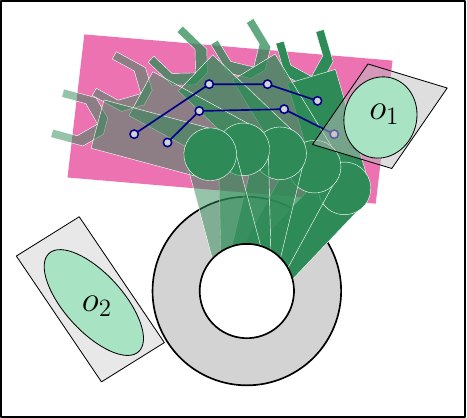}
        \label{fig:rgg:example:manip}
    }
    \hfill
    \subfloat[Mobile robot]{
        \includegraphics[width=0.45\linewidth]{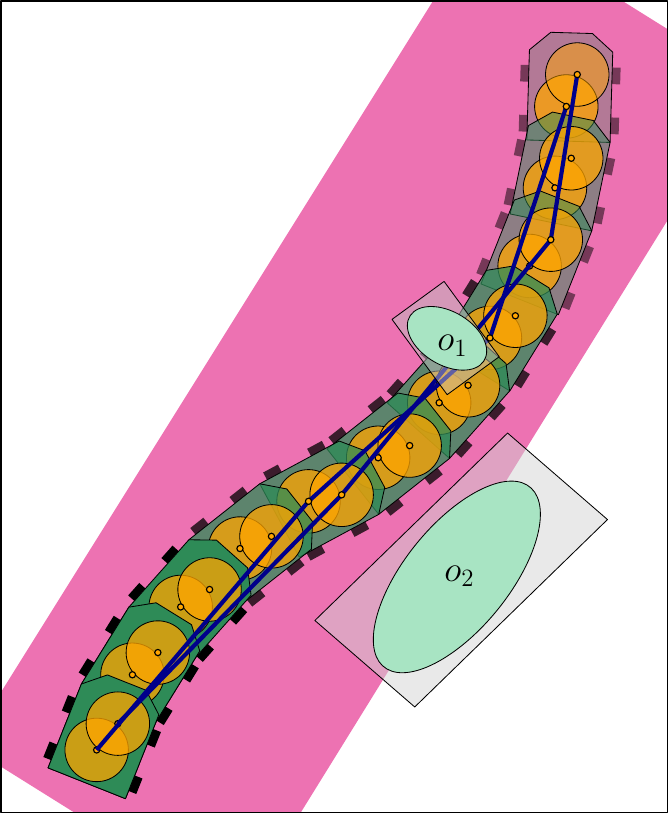}
        \label{fig:rgg:example:mobile}
    }
    \caption{Examples of outer- and inner-approximations of swept volumes corresponding to roadmap edges.
    A manipulator link and a mobile robot are over-approximated by oriented bounding boxes (pink) and under-approximated by splines (dark blue).
    Obstacles are over-approximated by OBBs (gray) and under-approximated by oval shapes (light blue).
    In (a), $o_1$ yields a gray edge (invalid) and $o_2$ yields a green edge.
    In (b), $o_1$ yields a gray edge (valid) and $o_2$ yields a red edge.}
    \label{fig:rgg:example}
\end{figure}


\section{Related Work}
\label{sec:related:work}
\subsection{Sampling-Based Motion Planning}
Roadmap graphs were popularized for robot motion planning with the introduction of the Probabilistic Roadmap algorithm (PRM) \cite{kslo-prpp-96}, which gave rise to a host of sampling-based motion planning (SBMP) algorithms. See ~\cite{ock-sbmpcr-23} and references therein for a comprehensive review of SBMP methods. Some SBMP algorithms, like PRM, construct a roadmap to efficiently answer multiple motion planning queries for a single robot in a fixed environment. Others, like LazyPRM \cite{bk-ppulp-00} and RRT \cite{l-rrtntpp-1998,kl-rrtceaspp-00}, are designed to answer a single query, making them useful building blocks for many algorithms used in dynamic settings.   


\subsection{Dynamic Roadmaps}
Dynamic roadmap graphs that can be updated as obstacles change position were introduced by Leven and Hutchinson \cite{lh-frtppce-02}. Their algorithm partitions the workspace using a grid, and maps each grid cell to the nodes and edges that intersect it, that is, those associated with robot configurations occupying some of the cell's volume. When an obstacle moves, only the roadmap components in the cells intersected by the obstacle's axis-aligned bounding box (AABB) are re-validated. This dynamic roadmap, commonly known as DRM, was later implemented and tested by Kallmann and Matari\`{c} \cite{km-mpdr-04}, who showed it is more efficient in answering motion planning queries in a modified environment than RRT. Yang \etal \cite{ymilv-hdrmrcdrrtmpcs-17} create a hierarchical DRM for articulated robots, which uses the hierarchy induced by the ordering of the bodies composing the robot to efficiently update sets of nodes and edges in a grid roadmap. The SPITE method \cite{almmmhpa-spitme-24}, which forms the basis for this work, sought to overcome the limitations of grid-based dynamic roadmaps by using different geometries. By approximating the swept volumes of the robot's motions using primitives known as cigars (or line-swept spheres) and storing them in an AABB tree for fast retrieval. This approach was shown to result in improved motion planning runtimes when compared to DRM, RRT, and LazyPRM. 

\subsection{Geometric Approximations and Collision Detection}
Proximity queries and collision detection methods are fundamental to motion planning \cite{jft-cdamp-05} and many other algorithmic fields, and have been widely studied \cite{lmk-cpq-17}. Various approaches have been proposed to compute or approximate geometric volumes, and swept volumes in particular \cite{amybj-swfpa-06}, which can be used accelerate collision detection \cite{lglm-pqp-99,lglm-fdqrssv-00}. \cite{dellin2016unifying} introduces a weighted Euclidian distance metric that can be calibrated to approximate swept volume by using mean swept volume per joint motion. More recent approaches \cite{chiang2018fast} use deep learning to speed up swept volume computations, though they introduce up to a 10\% error rate. 

\subsection{Serialization-based Acceleration in Motion Planning}
Hardware architectures that support data-parallel computation, such as GPUs and SIMD-enabled CPUs, have been utilized to accelerate motion planning algorithms. Most works focus on the collision detection aspect of motion planning which is both the bottleneck in many motion planning settings \cite{ksh-cdnnscbsbmp-16} and of a geometric nature, making serialization highly applicable. Pan and Manocha \cite{pm-gpubpcdfmp-12} gave a standalone GPU-based collision detection algorithm. Bialkowski et al. \cite{bkf-mprrtrrts-11} designed massively parallelized versions of RRT and RRT*, and Thomason et al. \cite{thomason2019motions} approximated the robot and workspace objects with spheres and used SIMD operations, both focused on collision checking acceleration and achieved significant speed ups over non-parallelized motion planning algorithms. Le et al. \cite{lpcwubcp-gtmp-25} recently presented a fully vectorized motion planning algorithm. In preliminary comparisons with VAMP, we encountered practical limitations in environments requiring large numbers of spheres for accurate geometric representations.



\section{Preliminaries}
\label{sec:preliminaries}

\subsection{Motion Planning Concepts and Notation}

 A motion planning problem consists of an environment (workspace) $W$ containing a set of obstacles $O$, a robot $r$ and start and goal configurations, $s$ and $t$, where the aim is to find a valid path for $r$ from configuration $s$ to configuration $t$. Let $R=(V,E)$ be a roadmap graph in the implicit configuration space $C_{space}$ induced by $r$ operating in $W$. We refer to the set $V \cup E$, consisting of all nodes and edges of $R$, as the \emph{roadmap components}. This unified view is justified by the fact that nodes may be treated as degenerate edges in our framework. 

\subsection{Swept Volumes and Approximations}

While roadmap edges are continuous curves in $C_{space}$, most algorithms approximate them into discrete sequences of configurations. We assume that every edge $e\in E$ is composed of $\frac{||e||}{\eps}$ configurations, where $||e||$ is the length of the edge, and $\eps > 0$ is the resolution parameter, and denote the set $\cfg(e)$.

The \emph{forward kinematics} function
\begin{equation*}
    \FK:C_{space} \longrightarrow P(\R^3)
\end{equation*}
maps sequences of \DOF{}s ($C_{space}$ coordinates) to configurations of $r$ in $W$. With some abuse of notation, we also use $\FK$ to mean its natural extension for sets $e\subseteq C_{space}$,

\begin{align*}
        \FK:(e) = \bigcup_{c\in \cfg(e)} \FK(c),
\end{align*}

and also the mapping of coordinates of objects from $O$ to their 3D volumes. We refer to the workspace volume produced by the forward kinematics mapping as a \emph{swept volume}, even when the input is a single configuration in $C_{space}$. Throughout the paper we discuss two types of approximations of geometries, called outer- and inner- approximations.  An outer-approximation (inner-approximation) of a compact set $C\subseteq \R^3$ is a superset (subset) of $C$, that is, a set $C'$  such that $C\subseteq C'$ ($C'\subseteq C$).

\section{Method}
\label{sec:method}

This section details the RGG framework and its vectorized implementation, Serialized RGG (SerRGG). We begin by introducing the RGG framework, designed to identify and label graph components as either invalid or valid (or of unknown validity) using heuristic geometric subroutines, thereby avoiding redundant, fine-grained collision checks. To support this, we describe our methods for approximating the robot's swept volumes and workspace obstacles using geometric primitives that are both computationally efficient and compatible with vectorized computations. Finally, we explain how these approximations are serialized to perform optimized, massively parallelized intersection checks on the GPU, enabling fast roadmap updates.

\subsection{The Red-Green-Gray Framework}
RGG uses two types of geometric heuristics to quickly determine the validity of a subset of a roadmap graph's nodes and edges. Both subroutines use an approximation of the swept volumes of roadmap components and obstacles, which are created during preprocessing and stored in AABB Trees. One subroutine checks for intersections (or lack thereof) between outer-approximations of both the obstacles and swept volumes of the robot, and the other performs similar checks for inner-approximations.

When an obstacle moves, its approximations are collision checked against the approximations of roadmap components. An intersection of the inner-approximations of the obstacle and component means the component is invalid (i.e., red), while no intersection between their outer-approximations means it is valid (i.e., green). If the outer-approximations intersect but the inner-approximations do not, the edge is labeled as having unknown validity (gray). See Figure \ref{fig:rgg:example} for examples. This laziness avoids unnecessary fine-grained (and computationally expensive) collision checking if the obstacle continues its motion or another obstacle moves in to invalidate the component before its validity needs to be assessed, say, for a lazy query. 

\subsection{Swept Volume Approximations}
Our choices of approximation methods were driven by three different optimization objectives. The first one is choosing approximations that can be easily serialized. The other objectives are as follows: Let $x \in V \cup E$ be a component of $R$ with outer-approximation $x^+$ and inner-approximation $x^-$, and let $o$ be an obstacle in $W$ with outer-approximation $o^+$ and inner-approximation $o^-$. 

For every such $x$ and $o$, we want to minimize the time spent on the collision detection/proximity queries $(x^+,o^+)$ and $(x^-,o^-)$, while minimizing
\begin{gather*}
    |\1[x^+ \cap o^+ \neq \emptyset] - \1[\FK(x) \cap \FK(o) \neq \emptyset]|%
    \\%
    \text{and }%
    \\%
    |\1[x^- \cap o^- \neq \emptyset] - \1[\FK(x) \cap \FK(o) \neq \emptyset]|,
\end{gather*}
where $\1[X]$ is the indicator function of event $X$. These latter terms are formal representations of the objective of capturing intersections of the underlying volumes using their approximations.

While SPITE uses line-swept spheres, also known as cigars or capped cylinders, to over-approximate the swept volume of nodes and edges, here we use an Oriented Bounding Box (OBB) to give a tighter approximation. For each body $D_i$ of the robot, we construct a point cloud $P$ from its intermediates as in SPITE and approximate the optimal volume OBB \cite{bhp-eamvbbpstd-01}. These bounding boxes are stored in an AABB tree, further described in \cite{almmmhpa-spitme-24}. All obstacles are also over-approximated with an OBB.

For each edge $e$, $k$ splines are constructed to under-approximate the swept volume. Splines are a simple geometry that can capture the "skeleton" of a robot's motion, and intersection tests with splines are significantly faster than volumetric mesh intersections. Each body $D_b$ of the robot is under-approximated using a set of spheres $S=\{s_i = (c_i,r_i) | 1\leq i \leq k\}$ such that $\bigcup_{i=1}^k s_i \subseteq D_b$ which can be placed manually or algorithmically \cite{sks-mssa-11}. The spheres are transformed with $D_b$ to the intermediate configurations along $e$ and the centers of copies of every sphere $s_i$ are sequentially connected, forming a spline $L_{e,i}$.

The complexity of $L_{e,i}$ depends on the length of $e$ and the resolution of the roadmap. Dense discretization or long edges result in high-complexity splines that significantly increase the computational cost of intersection tests. We therefore use a curve simplification algorithm by Agarwal \etal \cite{ahmw-nltaa-05} to find ``shortcuts'' that maintain the under-approximation property. For two indices $j<p$ in the spline, consider the line segment $l = c_{i,j}c_{i,p}$ between two copies of $s_i$ in the sequence associated with $e$. Intermediate indices $m$, such that $j<m<p$, can be removed from the spline if $l$ satisfies the following criterion:
\begin{equation*}
    \underset{j<m<p}{\max}\left(\text{dist}(l,c_{i,m})\right) < r_i
\end{equation*}
The shortcutted splines $L_e = \{L_{e,i} | 1 \leq i \leq k \}$ for all the robot bodies form the under approximation for the swept volume along an edge. As with the over-approximations, all of the splines are stored in an AABB tree. All obstacles are under-approximated with spheres.

\begin{figure}[t!]
    \centering
    \begin{subfigure}[t]{0.3\linewidth}
        \includegraphics[width=\linewidth,page=2]{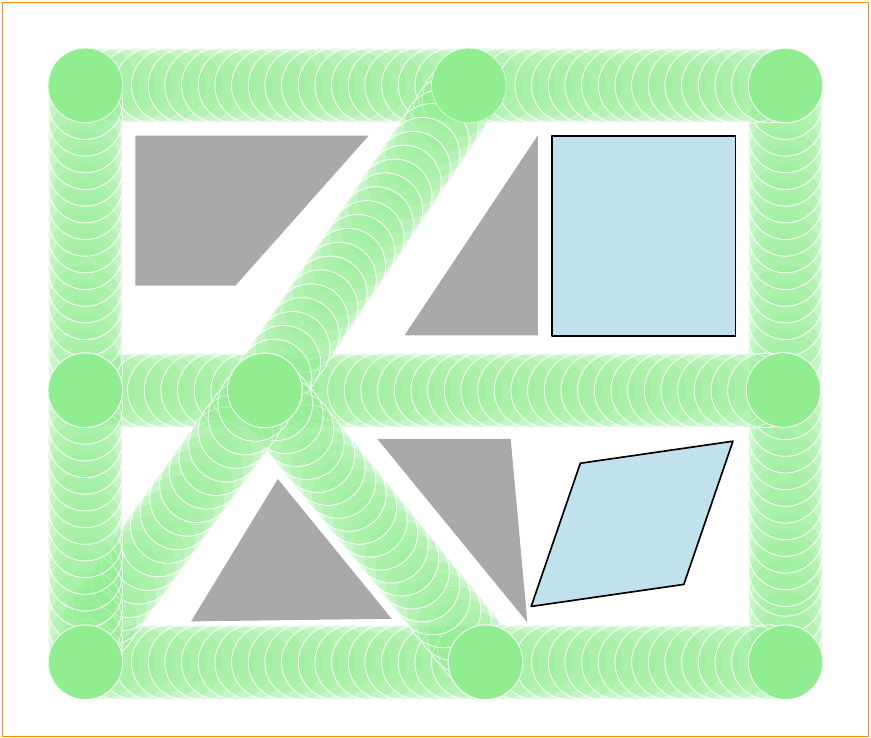}
        \caption{}
    \end{subfigure}
    \hfill
    \begin{subfigure}[t]{0.3\linewidth}
        \includegraphics[width=\linewidth,page=3]{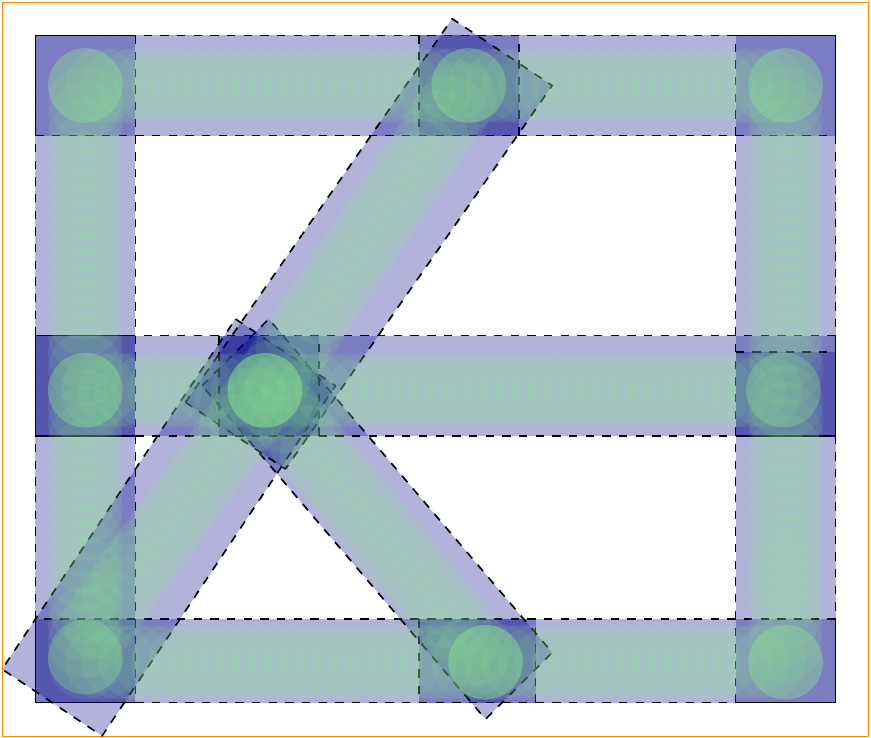}
        \caption{}
    \end{subfigure}
    \hfill
    \begin{subfigure}[t]{0.3\linewidth}
        \includegraphics[width=\linewidth,page=4]{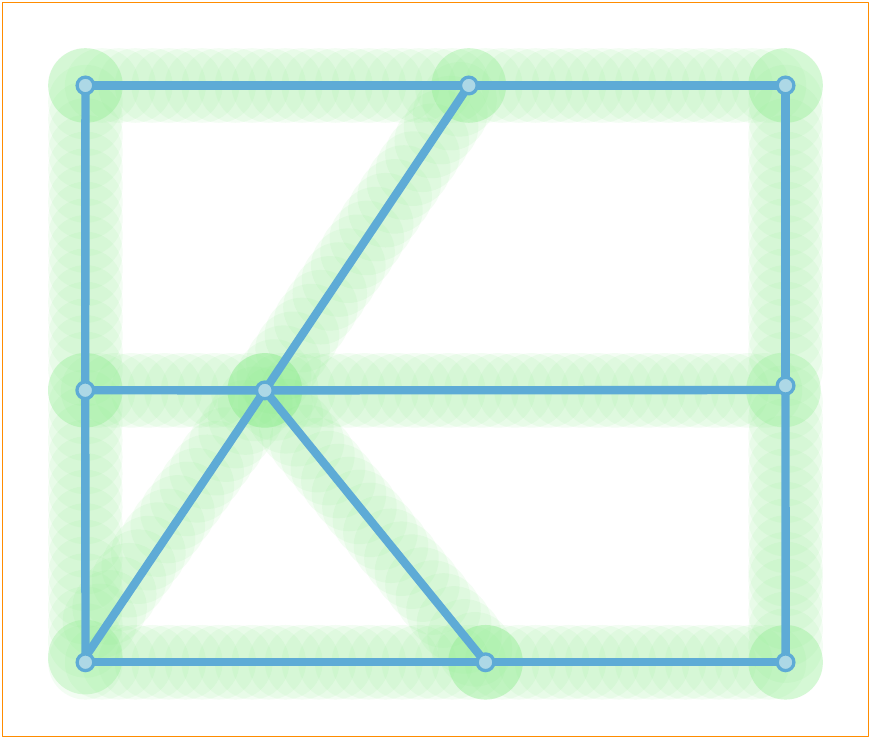}
        \caption{}
    \end{subfigure}
    \caption{Gray objects denote static obstacles while light blue objects denote dynamic obstacles. 
    (a) Initial roadmap validation with the obstacle in its starting position.
    (b) Over-approximations of the edges represented as OBBs in dark blue.
    (c) Under-approximations of the edges represented as splines in light blue.}
    \label{fig:serrgg2}
\end{figure}

\begin{figure}[t!]
    \centering
    \begin{subfigure}[t]{0.3\linewidth}
        \includegraphics[width=\linewidth,page=19]{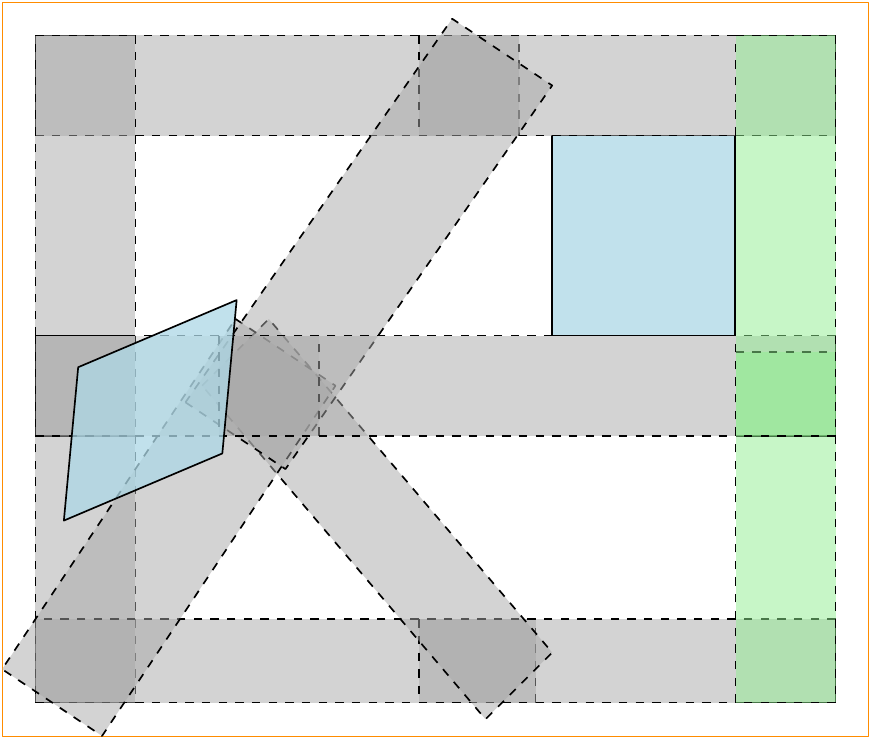}
        \caption{}
    \end{subfigure}
    \hfill
    \begin{subfigure}[t]{0.3\linewidth}
        \includegraphics[width=\linewidth,page=12]{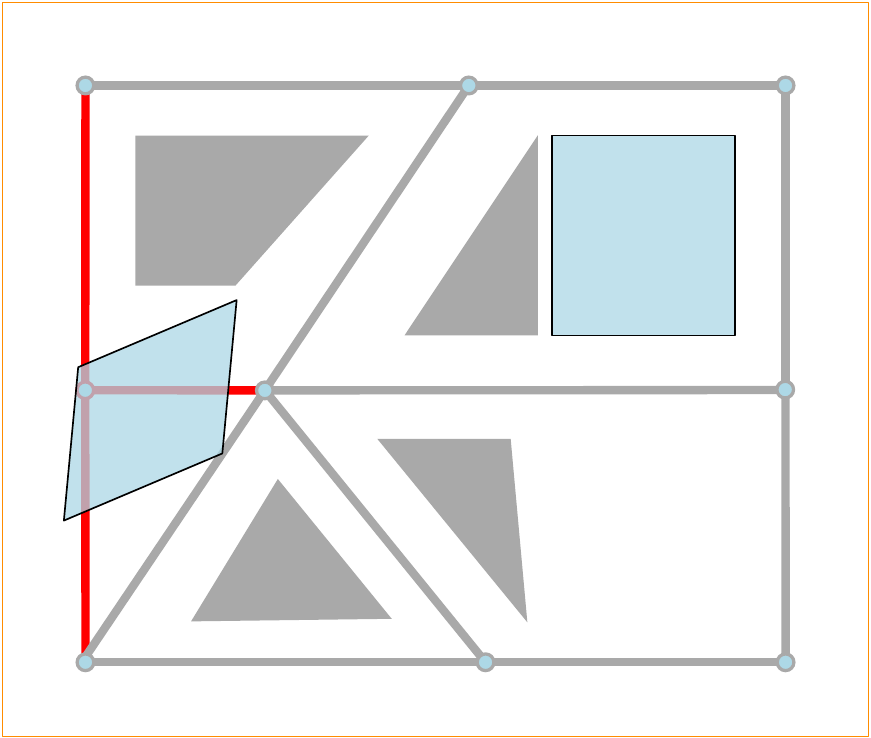}
        \caption{}
    \end{subfigure}
    \hfill
    \begin{subfigure}[t]{0.3\linewidth}
        \includegraphics[width=\linewidth,page=15]{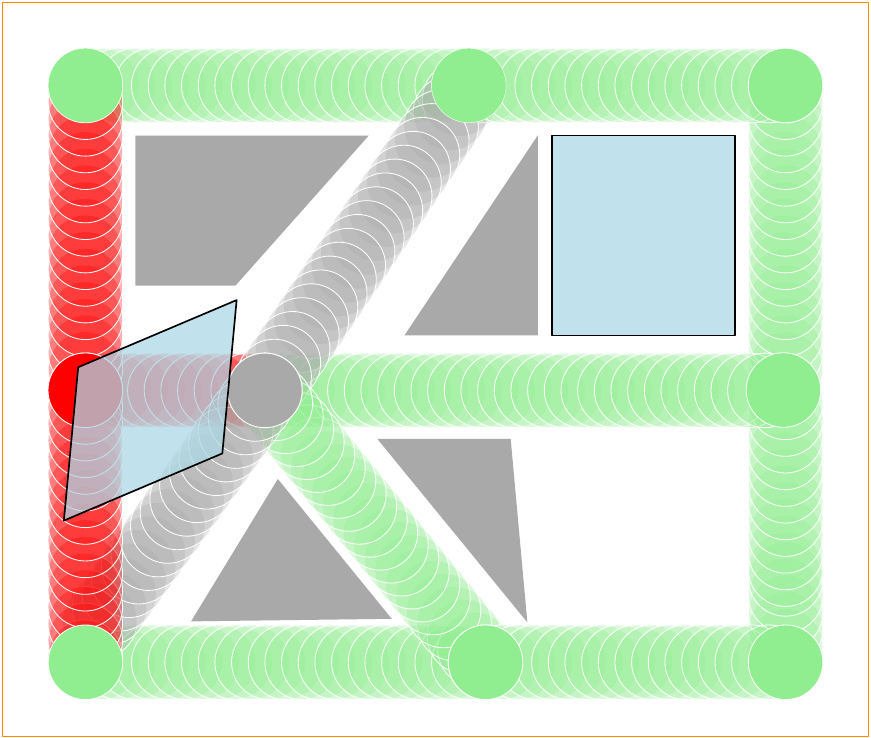}
        \caption{}
    \end{subfigure}
    \caption{Light blue objects denote dynamic obstacles. 
    (a) The edges corresponding to gray OBBs will be processed by SerRGG. 
    (b) Red splines correspond to edges that are invalid, whereas gray splines are unknown
    (c) Final edge classification produced by SerRGG, where red edges are invalid, gray edges remain unknown, and green edges are valid.}
    \label{fig:f4}
\end{figure}

\begin{algorithm}
    \caption{Roadmap Update with SPITE RGG}
    \label{alg:update_rm}
    \SetAlgoLined
    \SetKwInOut{Input}{input}
    \Input{Obstacle $o$, Transformation $t$, Boolean $lazy$}
    RevalidateOldIntersections()\\
    $o$.Update{}Location($t$)\\
    $b \gets AABB\_Over(o)$\\
    $invOverC \gets Tree\_Over$.Get{Intersecting}Objects($b$)\\
    \For{$c \in invOverC$}{
        \If{$c$.IsValid()}{
            $c$.MarkUnknown()\Comment{Edge over intersections marked unknown}\\
        }
        $c$.intersectionList.add($o$)\\
    }
    $s \gets AABB\_Under(o)$\\
    $invUnderC \gets Tree\_Under$.Get{Intersecting}Objects($s$)\\
    \For{$c \in invUnderC$}{
      $c$.MarkInvalid()\Comment{Edge under intersections marked invalid}\\
    }
    \If{not lazy}{
        \For{$c \in invOverC$}{
            \If{$c$.IsUnknown()} {
                \If{CD($c$,$o$) == free and not IsIntersectedByObstacles($c$)} {
                    $c.MarkValid()$
                }
                \Else {
                    $c.MarkInvalid()$
                }
            }
        }
    }
\end{algorithm}

The over- and under- approximations are used together to form the RGG paradigm or roadmap updates, summarized in Algorithm \ref{alg:update_rm}. Figure~\ref{fig:serrgg2} depicts the different approximations for the given roadmap. Given an obstacle $o$ whose pose changes by transformation $t$, the over-approximation AABB tree is queried with the obstacle's new OBB denoted $b$. For all nodes and edges with OBBs that intersect $b$, we mark them as unknown (gray). Next we query the under-approximation AABB tree with the obstacle's spheres, and set any intersecting nodes or edges to invalid (red). See Figure~\ref{fig:f4}. If not running the lazy version of the algorithm, any edges remaining gray from the first step have their validity determined with a full collision check. If running the lazy version, the edge remains gray and the determination of its validity is left for query time.

\subsection{Serial RGG}
To improve the computational speed of RGG updates, we present Serialized RGG (SerRGG). SerRGG leverages vectorized operations on the GPU to accelerate the heuristic collision checking of RGG.

Each point in the workspace is represented as a three-dimensional vector in $\mathbb{R}^3$. Each OBB is represented by its $8$ corner points and thus stored in a $8\times 3 $ matrix. Splines are stored as a sequence of line segments, where each segment is defined by two endpoints in $\mathbb{R}^3$, hence a $2\times 3$ array. For a spline with up to $K$ segments, we allocate a $K\times 2\times 3$ block to hold all endpoints. This uniform allocation ensures that all splines share the same memory layout, even when individual splines contain fewer than $K$ segments. 

Formally, we maintain the following fixed-dimension matrix stored as a PyTorch Tensor:
\begin{itemize}
    \item $\mathbf{E}^{+} \in \mathbb{R}^{N \times B \times 8 \times 3}$, storing the oriented bounding boxes (OBBs) of each edge for each robot body (outer approximations).
    \item $\mathbf{O}^{+} \in \mathbb{R}^{M \times 8 \times 3}$, storing OBBs of obstacles (outer approximations).
    \item $\mathbf{E}^{-} \in \mathbb{R}^{N \times B \times S \times K \times 2 \times 3}$, storing spline segments derived from medial-sphere approximations of edges (inner approximations).
    \item $\mathbf{O}^{-}_{c} \in \mathbb{R}^{M \times C \times 3}$ and $\mathbf{O}^{-}_{r} \in \mathbb{R}^{M \times 1 \times 1}$, storing the centers and radii of obstacle spheres (inner approximations).
\end{itemize}

Here $N$ is the number of edges in the roadmap, $M$ the number of obstacles, $B$ the number of robot bodies, $S$ the number of splines per body, $K$ the maximum number of segments used to represent each spline, and $C$ the number of spheres per obstacle.

\textbf{Offline storage.}
All of these matrices are constructed once in a canonical reference pose and stored persistently on the GPU. This avoids repeated data transfers and ensures coalesced memory access during queries.

\textbf{Online updates.}
When obstacles move, we do not rebuild the geometries. Instead, only the new transformations are uploaded to the GPU. The corresponding rows of the obstacle matrices ($\mathbf{O}^{+}, \mathbf{O}^{-}_{c}$) are updated in place by multiplying with the transformation matrices, while radii $\mathbf{O}^{-}_{r}$ remain unchanged. Again, this is done in order to avoid repeated data transfers. 

\textbf{Intersection queries.}
Once obstacle poses are updated, intersection kernels are launched over the serialized matrices, performing batched geometric intersection tests between roadmap elements and obstacles in parallel on the GPU.
\begin{itemize}
    \item OBB–OBB queries between $\mathbf{E}^{+}$ and $\mathbf{O}^{+}$ to certify edges as \emph{green} (valid).
    \item Spline–Sphere queries between $\mathbf{E}^{-}$ and $(\mathbf{O}^{-}_{c}, \mathbf{O}^{-}_{r})$ to certify edges as \emph{red} (invalid).
    \item Remaining edges are labeled \emph{gray}.
\end{itemize}
All operations are vectorized across the $N$ edges and $M$ obstacles, exploiting the serialized format for maximum GPU parallelism.

\textbf{Serialized OBB collision detection.}

The Separating Axis Theorem (SAT) is the standard approach for detecting intersections between oriented bounding boxes (OBBs) and other convex polytopes \cite{OBBSAT}. SAT leverages the fundamental principle that two convex objects are disjoint if and only if there exists a separating plane between them. For OBBs, this reduces to testing projections onto 15 specific axes: the face normals of both boxes (6 axes) and cross-products of their edge directions (9 axes) \cite{obbsatbook}.
SAT-based testing operates directly on the stored geometric representation, avoiding per-query transformations and yielding substantially faster validation when geometric data are reused across iterations. Therefore, our serialized implementation adopts the SAT-based approach, as it provides superior runtime performance while maintaining accuracy. 

\textbf{Spatially filtered many-to-many checking.}
In principle, validating a roadmap in a dynamic environment requires testing roadmap edges against nearby obstacles. 
A naive approach would perform an all-pairs check between $N$ edges and $M$ obstacles, resulting in up to $N\times M$ intersection tests per update. 
While highly parallel, this strategy quickly becomes wasteful when most edge-obstacle pairs are spatially irrelevant. 

To reduce unnecessary work, we introduce a lightweight one-layer grid data structure that spatially partitions the workspace. 
Each grid cell is sized to contain approximately the maximum number of primitives that can be efficiently processed by a GPU kernel without oversubscription. 
Roadmap edges are assigned to grid cells based on their spatial extent, and when obstacles move they are tested for intersection against grid cells. If an obstacle intersects a cell, only the edges contained in that cell are scheduled for detailed intersection testing.

In our GPU implementation, this results in a filtered many-to-many structure. 
Kernels are launched only for $(e,o)$ pairs where edge $e$ resides in a grid cell intersected by obstacle $o$. 
Within each kernel, we perform the relevant OBB-OBB and spline-sphere checks in parallel. 
This preserves the brute-force parallelism within each cell, while avoiding global all-pairs enumeration. 

\begin{figure}[t!]
    \centering
    \begin{subfigure}[t]{0.3\linewidth}
        \includegraphics[width=\linewidth,page=8]{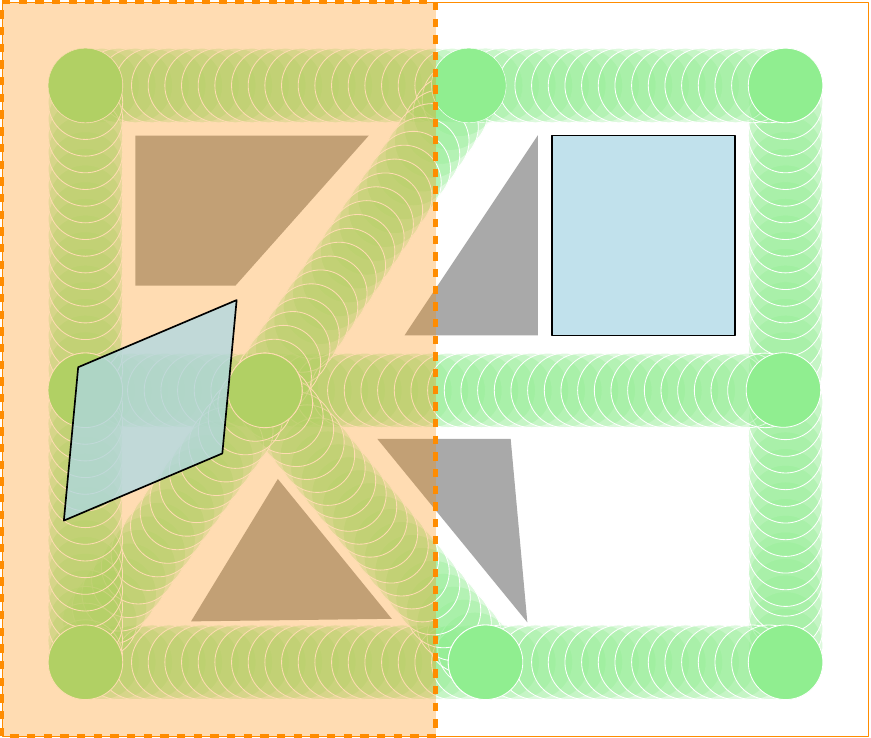}
        \caption{}
    \end{subfigure}
    \hfill
    \begin{subfigure}[t]{0.3\linewidth}
        \includegraphics[width=\linewidth,page=10]{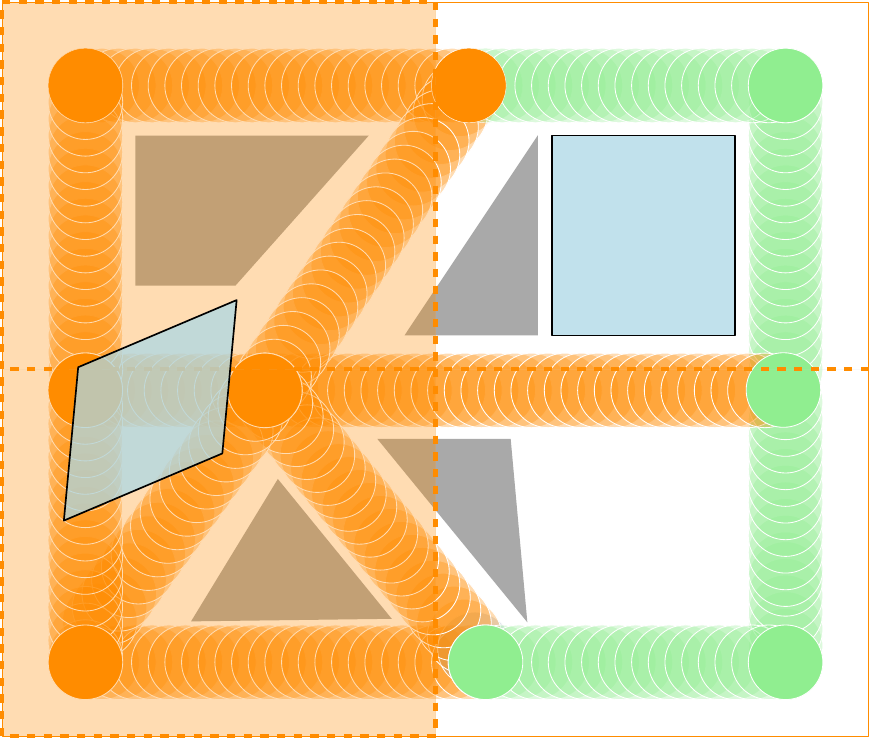}
        \caption{}
    \end{subfigure}
    \hfill
    \begin{subfigure}[t]{0.3\linewidth}
        \includegraphics[width=\linewidth,page=15]{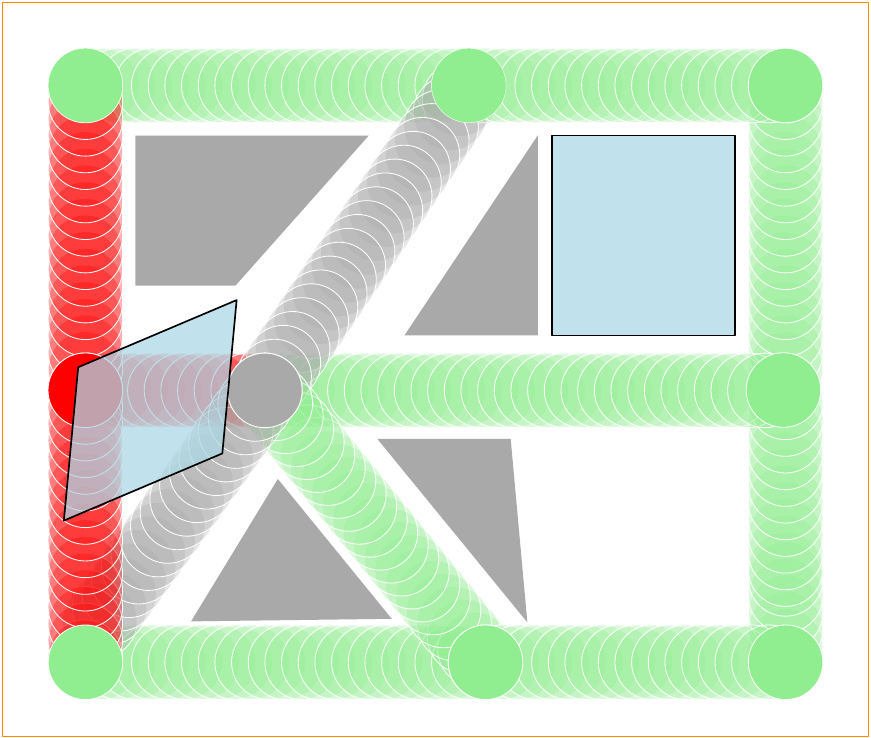}
        \caption{}
    \end{subfigure}
    \caption{Gray objects denote static obstacles while light blue objects denote dynamic obstacles. 
    (a) The region of the spatial grid affected by the obstacle's motion is highlighted in orange.
    (b) Roadmap edges associated with these grid cells are highlighted in orange and processed by SerRGG.
    (c) Final edge classification produced by SerRGG, where ref edges are invalid, gray edges remain unknown, and green edges are valid.}
    \label{fig:serrgg}
\end{figure}

This design has two advantages. First, the regular matrix representation of geometric approximations (OBBs as $8\times3$, splines as $K\times2\times3$) remains unchanged, allowing efficient vectorized GPU execution. Second, by spatially filtering interactions and restricting updates to cells affected by moving obstacles, we significantly reduce the number of intersection tests performed per update. 
As a result, our method balances aggressive parallelism with spatial locality, limiting computation to regions of the roadmap that are truly affected by dynamic changes. Our spatial filtering strategy is inspired by grid-based methods used in DRM, and by the AABB-tree-based filtering employed in SPITE, and is used here primarily for efficient GPU workload partitioning rather than to replace or approximate swept volumes. Figure~\ref{fig:serrgg} illustrates how obstacle motion activates a subset of grid cells and corresponding roadmap edges.

\textbf{Key advantage.}
By storing both edge and obstacle approximations in dense matrix form, and updating only the transforms of moved obstacles, SerRGG achieves millisecond-scale updates. Our method reduces the number of “unknown” edges while retaining the conservative guarantees required for safety.





\section{Experiments}
\label{sec:experiment}

We evaluate the roadmap update performance of RGG and SerGG against the original SPITE approach (outer-approximations only) for a simple rigid body robot and a 6-body articulated manipulator.
In each scenario, a probabilistic roadmap (PRM) is constructed offline and the outer and inner-approximations are generated to support RGG, SerRGG, and SPITE.
A sequence of random obstacle translation are performed, each triggering an update operation on the environment.
We report the time taken to perform the heuristic validation, the remaining unknown validities, and the time taken to compute those remaining validities.

For both robot types, we vary the size of the roadmaps to show how the update times scale with roadmap size.
For the mobile robot scenario, we also consider roadmap density along with the size and the number of objects being translated between updates.

\textbf{Experimental setup.} All experiments were run on a desktop computer with an Intel Core i7-14700F processor and an NVIDIA GeForce RTX 4060 GPU with 8 GB of memory. The Parasol Planning Library (PPL \cite{open-ppl}) implementations were used for all motion planning functions and algorithms.

\subsection{Mobile Robot Experiments}

We consider a simple unit cube robot navigating a large, open three-dimensional workspace of size 20x20x20.
The initial PRM is always built without obstacles in the environment.
The preprocessing time to generate the RGG geometries for each roadmap size is shown in Table \ref{tab:preprocessing_time}.





\begin{table}[t]
\centering
\caption{Preprocessing Time as a Function of Roadmap Size}
\label{tab:preprocessing_time}

\begin{tabular}{|r|r|r|r|r|r|r|r|r|r|}
\hline
\rowcolor{white}
\makecell{Nodes} & 10 & 100 & 1000 & 5000 & 10000 & 10000 & 20000 & 40000 & 60000 \\
\hline
\rowcolor{white}
\makecell{Edges} & 79 & 1204 & 11624 & 57148 & 115216 & 115069 & 230436 & 461054 & 691736 \\
\hline
\rowcolor{white}
\makecell{Preprocessing\\Time (s)} & 1.05 & 6.36 & 24.33 & 150.08 & 129.31 & 259.54 & 589.13 & 1087.74 & 1672.04 \\
\hline
\end{tabular}
\end{table}

The roadmap sizes for the experiment are scaled to observe the impact of the roadmap size on RGG and SerRGG, varying node numbers from 10 to 10k.
As the roadmap size increases while the environment size remains constant, the roadmap density (number of nodes and edges in a confined area) increases, resulting in more collisions with roadmap components when using the same size obstacles.
Because we also scale the obstacle, we can see the independent impact of roadmap density and obstacle size (Table \ref{tab:density_20_extended}).
We see that SerRGG generally stays around 800 $\mu$ s until the roadmap becomes sufficiently large at 10k nodes and it increases to around 2ms. RGG and SPITE scale linearly with roadmap size anywhere from 13 $\mu$ s to to 17 ms for RGG and 10 $\mu$ s to to 114 ms for SPITE. The number of unknown edges also scale linearly with the roadmap size, with RGG showing from 12\% to 30\% reduction in unknown edges as compared to SPITE (Table \ref{tab:density_20_extended}).

To isolate the effect of roadmap size independently of obstacle geometry, we perform a roadmap size experiment with a fixed obstacle size of $10\times 1\times1$ (Table~\ref{tab:roadmap_size}). In this setting, the number of roadmap nodes ranges from 5000 to 60000. As the roadmap size increases, the environment dimensions are scaled proportionally to maintain comparable roadmap density, this is reported as "factor" in Table~\ref{tab:roadmap_size}. We see that SerRGG grows about $200\mu $s with every increase of roadmap size, whereas RGG grows by about $2000\mu$s for the first three columns and then plateus around $20000$ with an environment of size $20\times20\times20$.

The study of the impact of the number of obstacles and the size of the obstacles is reported in Table~\ref{tab:num_obstacles}.
This was tested with both $3$ obstacles and $5$ obstacles.
Each obstacle is the same size and is translated to a new location at each iteration.
Roadmaps with $1000$ and $10000$ nodes were evaluated, along with obstacles of size $2\times2\times2$, $10\times1\times1$ and $10\times2\times 2$.
SerRGG takes around $1000\mu$s for every $1000$ node experiment, but grows as the size and density of the roadmap grows.
This is due the roadmap outgrowing the capacity of a single GPU instruction within the spatial grid regions.
Even with this, we still see up to 8x speed up.

\begin{table}[t]
\centering
\caption{Comparing SPITE, RGG and SerRGG for Environment Size $20\times20\times20$}
\label{tab:density_20_extended}

\rowcolors{2}{rowgray}{white}
\resizebox{\textwidth}{!}{
\begin{tabular}{|r|r|c|r|r|r|r|r|r|r|r|}
\hline
\makecell{Nodes} &
\makecell{Edges} &
\makecell{Obstacle\\Size} &
\makecell{SerRGG\\Time ($\mu$s)} &
\makecell{RGG\\Time ($\mu$s)} &
\makecell{SerRGG OA\\Time ($\mu$s)} &
\makecell{SPITE\\Time ($\mu$s)} &
\makecell{Unknown\\Edges (RGG)} &
\makecell{Validate RGG\\Time ($\mu$s)} &
\makecell{Unknown\\Edges (OA)} &
\makecell{Validate OA\\Time ($\mu$s)} \\
\hline

10 & 79 & $2\times2\times2$ & 818.0 & 13.8 & 558.2 & 10.2 & 3.17 & 605.2 & 3.62 & 625.5 \\
100 & 1204 & $2\times2\times2$ & 756.9 & 45.3 & 610.8 & 24.9 & 10.36 & 1366.9 & 11.65 & 1370.2 \\
1000 & 11624 & $2\times2\times2$ & 731.0 & 178.3 & 571.3 & 108.0 & 78.22 & 5378.2 & 86.77 & 5339.2 \\
10000 & 115216 & $2\times2\times2$ & 825.4 & 1252.5 & 659.7 & 605.9 & 563.19 & 25056.7 & 634.22 & 24810.0 \\
\hline

10 & 79 & $10\times2\times2$ & 768.2 & 22.5 & 583.0 & 17.6 & 9.49 & 1591.9 & 13.88 & 1545.4 \\
100 & 1204 & $10\times2\times2$ & 829.1 & 124.5 & 664.7 & 122.3 & 46.88 & 5756.5 & 59.21 & 5704.8 \\
1000 & 11624 & $10\times2\times2$ & 791.1 & 704.1 & 601.3 & 533.6 & 284.69 & 18896.7 & 359.65 & 18965.6 \\
10000 & 115216 & $10\times2\times2$ & 1290.6 & 7174.4 & 1113.4 & 4222.5 & 2015.56 & 65134.7 & 2603.98 & 65660.6 \\
\hline

10 & 79 & $20\times2\times2$ & 742.5 & 27.2 & 572.0 & 22.2 & 12.58 & 1810.9 & 19.79 & 1854.6 \\
100 & 1204 & $20\times2\times2$ & 795.5 & 210.3 & 630.0 & 205.0 & 82.98 & 8279.3 & 111.79 & 8282.1 \\
1000 & 11624 & $20\times2\times2$ & 863.5 & 1489.1 & 688.3 & 1461.8 & 528.70 & 33436.2 & 675.01 & 35356.2 \\
10000 & 115216 & $20\times2\times2$ & 1979.1 & 17318.3 & 2034.3 & 13915.7 & 3881.39 & 111579.0 & 4954.85 & 113253.8 \\
\hline

\end{tabular}}
\end{table}

\begin{table}[t]
\centering
\caption{Roadmap Size Test Results (Obstacle Size $10\times1\times1$)}
\label{tab:roadmap_size}

\rowcolors{2}{rowgray}{white}
\resizebox{\textwidth}{!}{
\begin{tabular}{|r|r|c|r|r|r|r|r|r|c|r|}
\hline
\makecell{Nodes} &
\makecell{Edges} &
\makecell{Obstacle\\Size} &
\makecell{SerRGG\\Time ($\mu$s)} &
\makecell{RGG\\Time ($\mu$s)} &
\makecell{Speedup} &
\makecell{Brute Force\\Time ($\mu$s)} &
\makecell{Unknown\\Fraction} &
\makecell{Validate Remaining\\Time ($\mu$s)} &
\makecell{Environment} &
\makecell{Factor} \\
\hline
5000 & 57148 & $10\times1\times1$ & 954.5 & 2396.1 & 2.5103 & 13502471.7 & 0.0198 & 80383.6 & $5\times20\times20$ & 1 \\
10000 & 115069 & $10\times1\times1$ & 1277.4 & 5423.8 & 4.2460 & 23635076.6 & 0.0179 & 125486.8 & $10\times20\times20$ & 2 \\
20000 & 230436 & $10\times1\times1$ & 1491.8 & 7220.9 & 4.8404 & 42407585.8 & 0.0105 & 142604.1 & $20\times20\times20$ & 2 \\
40000 & 461054 & $10\times1\times1$ & 1788.9 & 7207.1 & 4.0288 & 83371215.6 & 0.0030 & 87439.3 & $40\times20\times20$ & 2 \\
60000 & 691736 & $10\times1\times1$ & 2042.5 & 7315.1 & 3.5814 & 121769499 & 0.0034 & 166644.5 & $60\times20\times20$ & 1.5 \\
\hline
\end{tabular}}
\end{table}

\begin{table}[t]
\centering
\caption{Number of Obstacles Test Results (Environment $20\times20\times20$)}
\label{tab:num_obstacles}

\rowcolors{2}{rowgray}{white}
\resizebox{\textwidth}{!}{
\begin{tabular}{|r|r|r|c|r|r|r|r|r|r|}
\hline
\makecell{Nodes} &
\makecell{Edges} &
\makecell{\# Obstacles} &
\makecell{Obstacle\\Size} &
\makecell{SerRGG\\Time ($\mu$s)} &
\makecell{RGG\\Time ($\mu$s)} &
\makecell{Speedup} &
\makecell{Brute Force\\Time ($\mu$s)} &
\makecell{Unknown\\Fraction} &
\makecell{Validate Remaining\\Time ($\mu$s)} \\
\hline
1000 & 11624 & 3 & $2\times2\times2$ & 942 & 547 & 0.5807 & 8618093.6 & 0.0249 & 71081 \\
10000 & 115216 & 3 & $2\times2\times2$ & 1617.9 & 4284.7 & 2.6483 & 40917146 & 0.0177 & 203030 \\
\hline
1000 & 11624 & 3 & $10\times1\times1$ & 1020.7 & 1317 & 1.2903 & 8738171 & 0.0450 & 131055.6 \\
10000 & 115216 & 3 & $10\times1\times1$ & 2874.3 & 11920.3 & 4.1472 & 41164237.6 & 0.0311 & 364856.6 \\
\hline
1000 & 11624 & 3 & $10\times2\times2$ & 1069.3 & 2218 & 2.0743 & 8781372.3 & 0.0670 & 211223.7 \\
10000 & 115216 & 3 & $10\times2\times2$ & 3646.2 & 30734.4 & 8.4292 & 42120244.7 & 0.0497 & 599736.5 \\
\hline
1000 & 11624 & 5 & $2\times2\times2$ & 1019 & 929.1 & 0.9118 & 11961060 & 0.0379 & 136936.7 \\
10000 & 115216 & 5 & $2\times2\times2$ & 2679.9 & 8092.8 & 3.0198 & 55886801.6 & 0.0273 & 418541.5 \\
\hline
1000 & 11624 & 5 & $10\times1\times1$ & 1213.2 & 2234.7 & 1.8420 & 12002679.7 & 0.0680 & 253691.8 \\
10000 & 115216 & 5 & $10\times1\times1$ & 5501.9 & 22687.3 & 4.1235 & 56272723.3 & 0.0486 & 688820.5 \\
\hline
1000 & 11624 & 5 & $10\times2\times2$ & 1302.1 & 3946.9 & 3.0312 & 12040866.4 & 0.0987 & 389801.9 \\
10000 & 115216 & 5 & $10\times2\times2$ & 7015.1 & 62155.9 & 8.8603 & 58070418.9 & 0.0732 & 1068508.8 \\
\hline
\end{tabular}}
\end{table}

\subsection{Manipulator Experiments}
We evaluate RGG and SerRGG on a 6-body articulated manipulator, constructing the PRM and the outer and inner-approximations in a fixed 3D environment containing a single obstacle.
Results are reported in Table~\ref{tab:manipulator_test}.

We again see SerRGG begin to offer speedup in the update time once the roadmap increases in size with roadmaps of 1000 and 5000 nodes seeing a roughly 2x speedup (this trend is repeated with SerRGG OA and SPITE).
Comparing RGG to SPITE (and SerRGG to SerRGG OA), it is clear to see that the majority of the RGG update time is spent on the inner-approximation intersection checks.
When including the additional time spent validating the remaining unknown edges, the total time for RGG and SPITE is near even.
With the GPU acceleration of both variants, the total validation time when first applying SerRGG is less than SerRGG OA while producing a more informed roadmap after the heuristic validation (as seen in the roughly 15\% reduction in the number of unknown edges).

\begin{table}[t]
\centering
\caption{Manipulator Test Results}
\label{tab:manipulator_test}

\rowcolors{2}{rowgray}{white}
\resizebox{\textwidth}{!}{
\begin{tabular}{|r|r|c|r|r|r|r|r|r|r|r|}
\hline
\makecell{Nodes} &
\makecell{Edges} &
\makecell{Obstacle\\Size} &
\makecell{SerRGG\\Time ($\mu$s)} &
\makecell{RGG\\Time ($\mu$s)} &
\makecell{SerRGG OA\\Time ($\mu$s)} &
\makecell{SPITE\\Time ($\mu$s)} &
\makecell{Unknown\\Edges (RGG)} &
\makecell{Validate RGG\\Time ($\mu$s)} &
\makecell{Unknown\\Edges (OA)} &
\makecell{Validate OA\\Time ($\mu$s)} \\
\hline
100   & 255.4   & $3\times3\times3$ & 645.7   & 274.9   & 528.0   & 125.5   & 32.81   & 5478.2   & 38.35   & 5860.5  \\
1000  & 3269    & $3\times3\times3$ & 2062.7  & 3937.7  & 795.9   & 1341.8  & 300.20  & 42829.1  & 358.28  & 45816.0 \\
5000  & 227680  & $3\times3\times3$ & 12675.6 & 26256.6 & 4110.0  & 7121.0  & 1927.03 & 214432.6 & 2316.57 & 235806.1 \\
\hline
\end{tabular}}
\end{table}

\section{Discussion}
\label{sec:discussion}

Although under-approximations are not strictly required for correctness, our results demonstrate that they provide consistent benefits with minimal overhead. The additional computational cost of using the under-approximations is minimal compared to the reduction in time from preemptively identifying invalid edges. As seen in Table~\ref{tab:density_20_extended}, RGG reduces the burden of full collision checks, especially as the total number of potentially unknown edges increases.

RGG provides a substantial benefit in reducing the number of unknown edges needing full validation. In sparse roadmaps, the sequential RGG outperforms SerRGG as the GPU provides limited benefit because the parallelism is underutilized. However, as roadmap density increases, SerRGG demonstrates superior scalability. In particular, SerRGG is most effective with large obstacles or extensive changes in the environment. RGG update times grow with the number of invalidated edges due to its sequential validation process, whereas SerRGG exhibits a much slower growth rate by leveraging vectorized GPU execution. This is demonstrated in Table~\ref{tab:density_20_extended}, where SerRGG consistently outperforms the sequential implementation on larger roadmaps, achieving a speedup factor up to 8.75x in scenarios with large obstacles. 

On very small graphs, SerRGG is limited by GPU transfer overhead; however, once roadmaps reach sufficient scale, it outperforms RGG by an average factor of 4x (Table~\ref{tab:roadmap_size}). A similar performance gap is seen when introducing multiple obstacles (Table~\ref{tab:num_obstacles}). Since RGG performs edge validation sequentially, its runtime degrades rapidly as obstacle interactions increase, whereas SerRGG's growth rate is significantly slower. Manipulator experiments further validate this trend, demonstrating that SerRGG's speedup relative to RGG increases as the roadmap size grows (Table~\ref{tab:manipulator_test}).

\section{Conclusion}

In this paper, we introduced the Red-Green-Gray (RGG) framework and its serialized, GPU-oriented realization, SerRGG. Collectively, these results demonstrate the strengths of both RGG and SerRGG. The RGG framework introduces a structured edge classification scheme based on geometric over- and under-approximations, enabling selective updates and limiting expensive exact collision checks to ambiguous cases. Building on that geometric foundation, SerRGG fundamentally changes how revalidation scales as problem complexity increases. By shifting edge validation from a sequential CPU process to a vectorized, batched GPU operation, SerRGG transforms revalidation from a bottleneck into a workload that scales favorably with roadmap density and obstacle interaction. 

The consistently low fraction of unknown edges further suggests that shallow updates using only the approximations are sufficient to resolve most edge statuses in practice, limiting the need for expensive followup validation. As roadmap sizes, obstacle counts, and geometric complexity continue to grow in real-world applications these properties position SerRGG as a practical and scalable foundation for real-time roadmap revalidation, particularly in settings where frequent environmental changes make repeated full validation infeasible. Future work will include integrating SerRGG into full motion planning pipelines and evaluating its impact on end-to-end planning performance. 

\bibliographystyle{splncs04}
\bibliography{robotics.bib}


\end{document}